\documentclass{article}
\usepackage{spconf,amsmath,graphicx,hyperref}
\usepackage{multirow}
\usepackage{amssymb}
\usepackage{amsmath}

\title{Progressively Texture-Aware Diffusion for Contrast-Enhanced Sparse-View CT}
\name{Tianqi Wang \qquad Wenchao Du\sthanks{Corresponding author} \qquad Hongyu Yang}
\address{The College of Computer Science, Sichuan University, Chengdu 610065, China\\
wangtianqi2@stu.scu.edu.cn, \{wenchaodu.cs, yanghongyu\}@scu.edu.cn\\
}
\begin{document}
%
\maketitle
\begin{abstract}
Diffusion-based sparse-view CT (SVCT) imaging has achieved remarkable advancements in recent years, thanks to its more stable generative capability. However, recovering reliable image content and visually consistent textures is still a crucial challenge. In this paper, we present a Progressively Texture-aware Diffusion (PTD) model, a coarse-to-fine learning framework tailored for SVCT. Specifically, PTD comprises a basic reconstructive module PTD$_{\textit{rec}}$ and a conditional diffusion module PTD$_{\textit{diff}}$. PTD$_{\textit{rec}}$ first learns a deterministic mapping to recover the majority of the underlying low-frequency signals (i.e., coarse content with smoothed textures), which serves as the initial estimation to enable fidelity. Moreover, PTD$_{\textit{diff}}$ aims to reconstruct high-fidelity details for coarse prediction, which explores a dual-domain guided conditional diffusion to generate reliable and consistent textures. Extensive experiments on sparse-view CT reconstruction demonstrate that our PTD achieves superior performance in terms of structure similarity and visual appeal with only a few sampling steps, which mitigates the randomness inherent in general diffusion models and enables a better trade-off between visual quality and fidelity of high-frequency details.
\end{abstract}
\section{Introduction}
\label{sec:intro}
 X-ray Computed Tomography (CT) is a highly valuable and widely used medical imaging technology, yet it exposes patients to significant levels of radiation, potentially harming their health \cite{smith2009radiation}. While existing learning-based methods have achieved notable breakthroughs in mitigating latent risks by reducing radiation dose, they typically employ a mapping model to produce deterministic results. A persistent challenge is that these approaches frequently yield unsatisfactory details, as they are trained to minimize pixel-level error, aligning output with routine-dose CT images by Norm-based losses. Generative adversarial networks (GANs) have been exploited to generate visually pleasing results \cite{yang2018low}. However, undesired artifacts always emerge in the results due to their unstable adversarial nature.

Diffusion models (DMs) have revolutionized image generation and yielded diverse and realistic image content \cite{ho2020denoising}. Leading studies have explored conditional DMs for low-dose CT to produce perceptually appealing results that closely resemble routine-dose CT images \cite{gao2023corediff}. However, a crucial challenge is that recovering the high-fidelity content required in CT imaging clashes with the stochastic nature of diffusion models. A plausible solution involves integrating the physical degradation models with neural networks or crafting hand-designed priors to mitigate the inherent randomness of the diffusion paradigm. Although previous studies \cite{Song2022SolvingIP} have adopted this approach to ensure reliable structure restoration, they lack generalization ability and introduce undesired artifacts.

To address the above problems, we present a Progressively Texture-aware Diffusion model (shortened by PTD), a coarse-to-fine learning architecture, to reconstruct contrast-enhanced CT images. PTD integrates the idea of divide-and-conquer into an end-to-end training scheme, i.e., comprising a basic reconstructive module PTD$_{\textit{rec}}$ and a conditional diffusion module PTD$_{\textit{diff}}$. PTD$_{\textit{rec}}$ models a deterministic mapping to recover low-frequency image content (i.e., smoothed structures), which provides an initial estimation. By focusing on low-frequency content, PTD$_{\textit{rec}}$ ensures structural consistency and maintains fidelity, serving as a reliable starting point for the diffusion process. PTD$_{\textit{diff}}$ models a bijective mapping between low-frequency and high-frequency image domains, which explores the dual-domain conditional diffusion model to generate contrast-enhanced structure details. By incorporating low and high-frequency content priors into the diffusion network, PTD can further eliminate noise and artifacts while recovering more reliable details. Extensive experiments demonstrate that PTD achieves superior performance against state-of-the-art methods on structural preservation and visual perception. In short, our contributions are summarized: \textbf{1)} Propose a novel Progressively Texture-aware Diffusion framework for high-fidelity CT imaging, which integrates the idea of divide-and-conquer into the general diffusion process, recovering signals in a coarse-to-fine manner; \textbf{2)} Construct a dual-domain guided conditional diffusion model, which exploits image and multiscale feature prompts to drive reliable and consistent texture generation; \textbf{3)} Extensive experiments on sparse-view CT demonstrate that our method achieves a better trade-off between structural preservation and visual appeal.

\section{Method}
\subsection{Preliminaries of Diffusion Schrödinger Bridge}
To make the paper self-contained, we first briefly review the Diffusion Schrödinger Bridge (DSB) \cite{Liu2023I2SBIS}. DSB directly models a bijective translation between source and target image domains, and has presented promising progress in natural image restoration and low-dose CT reconstruction \cite{du2024structure}.

Assuming the paired data $(X_0, X_1)$ is available, $X_0 \in p_\mathcal{A}$ and $X_1 \in p_{\mathcal{B}}$ are drawn from boundary distributions in two distinct domains, $X_t$ is designed to follow the distribution:
\begin{equation}
	\begin{aligned}
		&q(X_t|X_0,X_1)= \\
		&\mathcal{N}(X_t; \frac{\bar{\sigma}_t^2}{\bar{\sigma}_t^2 + \sigma_t^2}X_0 + \frac{{\sigma}_t^2}{\bar{\sigma}_t^2 + \sigma_t^2}X_1; \frac{\sigma_t^2\bar{\sigma}_t^2}{\bar{\sigma}_t^2 + \sigma_t^2}I),
	\end{aligned}
	\label{eq1}
\end{equation}
where $t\in [0, 1]$ denotes the discrete time-step by $0=t_0 < \cdots t_n < \cdots < t_N = 1$, $\sigma^2_t = \int_{0}^{t}\beta_{\tau}\mathrm{d}\tau$ and $\bar{\sigma}^2_t = \int_{t}^{1}\beta_{\tau}\mathrm{d}\tau$ represent variances accumulated from either side, and $\beta$ determines the speed of diffusion. Assuming $X_t$ is sampled analytically using Eq~\ref{eq1}, the denoising network $\epsilon(X_t, t;\theta)$ can be efficiently trained to predict the difference between $X_t$ and $X_0$ by minimizing the loss function:
\begin{equation}
	\begin{aligned}
		&\theta^{\ast}=\arg\min_{\theta}\mathbb{E}_{X_0, X_1} \\
		&\mathbb{E}_{t\sim\mathcal{U}[0, 1],X_t\sim q(X_t|X_0,X_1)}\|\epsilon(X_t, t;\theta) - \frac{X_t-X_0}{\sigma_t}\|.
	\end{aligned}
	\label{eq2}
\end{equation}

\begin{figure}[t]
	\centering
	\includegraphics[width=1.\linewidth]{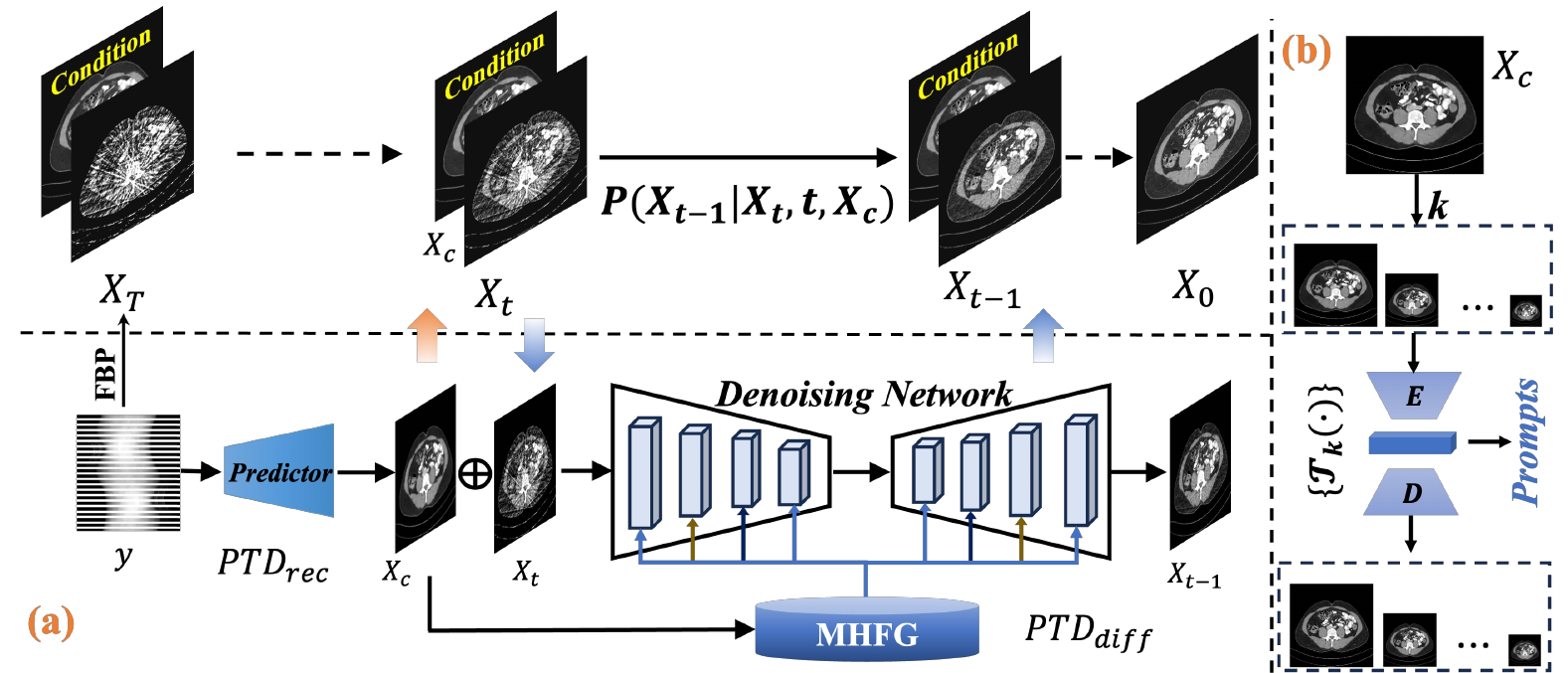}
	\caption{Overview of framework. (a) PTD first exploits a deterministic predictor to produce an initial results, and then which joint the Multiscale High-Frequency texture Guidance (MHFG) as extra prompts to guide high-frequency signal generation with a Diffusion Schrdinger Bridge (DSB) model; (b) MHFG module, it explores the multi-scale lightweight super-resolution networks to generate fine texture prompts.}
	\label{fig1}
\end{figure}

During the generative stage, DSB first begins with $X_N$ and iteratively approaches the target domain image $X_0$. In the step from $X_{n+1}$ to $X_n$, $\hat{X}_0$ is first calculated by the pretrained $\epsilon$ and $X_{n+1}$:
\begin{equation}
	\hat{X}_0 = X_{n+1} - \sigma_{n+1}\epsilon(X_{n+1}, t_{n+1}; \theta^{\ast}),
\end{equation}
where $\hat{X}_0$ is an approximation of $X_0$. $X_n$ is sampled from the DDPM \cite{ho2020denoising} posterior $p$ using $\hat{X}_0$ and $X_{n+1}$:
\begin{equation}
	X_n \sim p(X_n | \hat{X}_0, X_{n+1}),
	\label{eq4}
\end{equation}
here the posterior $p$ is expressed as
\begin{equation}
	\begin{aligned}
		&p(X_n|X_0,X_{n+1}) = \\ &\mathcal{N}(X_n;\frac{\alpha^2_n}{\alpha^2_n+\sigma^2_n}X_0 + \frac{\sigma^2_n}{\alpha^2_n+\sigma^2_n}X_{n+1}, \frac{\alpha^2_n\sigma^2_n}{\alpha^2_n+\sigma^2_n}I ),
	\end{aligned}
\end{equation}
where $\alpha^2_t = \int_{t_n}^{t_{n+1}}\beta_{\tau}\mathrm{d}\tau$ denotes the accumulated variance between consecutive time steps $t_n$ and $t_{n+1}$. The DDPM posterior $p$ satisfies the equation
\begin{equation}
	\begin{aligned}
		&q(X_n|X_0,X_N)= \\
		&\int p(X_n|X_0, X_{n+1})q(X_{n+1}|X_0,X_N)\mathrm{d}X_{n+1},
	\end{aligned}
\end{equation}
which guarantees that $X_n$ sampled using Eq~\ref{eq4} is exactly a sample from $q(X_n|X_0,X_N)$ assuming the trained network $\epsilon^{\ast}$ is perfect. In fact, training a perfect $\epsilon^{\ast}$ is difficult, which clashes with the stochastic nature of the diffusion process, and also requires extensive data and lengthy training cycles for effective learning. Therefore, PTD aims to explore a conditional DSB model to alleviate the stochastic nature of the diffusion, the whole framework is shown in Fig~\ref{fig1}.
\subsection{Dual-Domain Guided Conditional DSB}
\subsubsection{Initial Predictor} The basic reconstructive module PTD$_{\textit{rec}}$ aims to recover the coarse structure and content of the CT image from the low-dose data $y$. Following \cite{chen2017low,chen2018learn}, PTD$_{\textit{rec}}$ directly learns a deterministic predictor $F_{\theta}(\cdot)$ between the sparsely sampled data (i.e., $y$) and the corresponding routine-dose data $X_0$. Therefore, PTD$_{\textit{rec}}$ can be implemented by any existing CNN-based model, which would produce an initial estimation $X_{init}$. The content loss $\mathcal{L}_{Content}$ is defined as the mean squared error between the coarse prediction $X_{init}$ and $X_0$.

\subsubsection{Low-Frequency Guided DSB} PTD$_{\textit{diff}}$ aims to generate high-frequency details through the dual-domain conditional DSB, where the visual prompts from pixel and feature spaces are used to guide reliable texture generation. Therefore, the $X_{init}$ is directly used as a conditional input $X_c$ to guide the diffusion process. For the feature prompt, we further explore the multiscale texture representations to guide consistent high-frequencies generation, which serves as implicit bias that informs the PTD$_{\textit{diff}}$ about the high-frequency prior at the intermediate layers.

PTD$_{\textit{diff}}$ joint the sparse-view CT image (i.e., $X_T$) and initial estimation $X_{init}$ as input, bridging a domain translation between the noisy CT image $X_T$ and the routine-dose CT image $X_0$, so that the Eq~\ref{eq2} is rewritten as $\epsilon(X_t, t, X_{c}; \theta)$. Considering that the low-frequency image priors have been coupled into the denoising network, PTD$_{\textit{diff}}$ will focus on high-frequency detail restoration. We have noted that the recent advances \cite{ye2024learning} explored the residual diffusion paradigms to generate high-frequency details from noisy inputs. However, \textit{there is a key challenge lies in that high- and low-frequency signals are generally coupled in pixel space, directly generating consistent and reliable textures from noisy data remains challenging, which can not achieve a better trade-off between structure preservation and visual appeal}.

\subsubsection{Multiscale High-Frequency Guidance} In order to guide consistent and reliable texture generation, we further introduce a Multiscale High-Frequency Guidance module (MHFG, shown in Fig~\ref{fig1}(b)), which exploits the explicit texture representation as an initial high-frequency estimation, and guides high-fidelity diffusion. Specifically, MHFG explores the lightweight super-resolution network $\mathcal{T}(\cdot)$ to generate multiscale texture representations from conditional low-frequency $X_{c}$, therefore forming a specific super-resolution task to learn the latent texture prior distribution to some degree,
\begin{equation}
	\mathcal{L}^k_{\textit{Guidance}} = \mathbb{E}\left\| (\mathcal{T}_k(\downarrow_k(X_{init}))) -  \downarrow_k(X_0)\right\|_2
	\label{eq9}
\end{equation}
where $\downarrow_k(\cdot)$ denotes the downsampling operation with a factor of $2^k$. In practice, we set $k=1,2,3,4$, i.e., MHFG contains four light super-resolution subnetworks $\{\mathcal{T}_k(\cdot)\}$, each of which shares the same network but receives inputs of different scales. Therefore, Eq~\ref{eq9} is rewritten as $\mathcal{L}_{Guidance}=\sum_{k=1}^{K}\mathcal{L}^k_{Guidance}$. Following \cite{Dong2014ImageSU}, we implement each subnetwork with a lightweight three-layer CNNs, and the output of the intermediate layer is used as latent texture representation. As shown in Fig~\ref{fig1}(a), we insert the MHFG module into the denoising network $\epsilon^{\ast}$ by adding the extracted representations to the feature maps at the respective scale on the encoder and decoder as the specific biases. To compensate for the depth difference, as each corresponding scale, we apply a convolutional layer with $1\times 1$ kernel that has the same number of feature channels as those in the encoder and decoder.

By aggregating the information from the condition $X_{c}$, and the multiscale texture guidance $\{\mathcal{T}_k(X_{c})\}$, the denoising loss of the conditional DSB model from Eq~\ref{eq2} is rewritten as:
\begin{equation}
	\mathcal{L}_{\textit{Diff}} = \mathbb{E}\|\epsilon(X_t, t, X_{c}, \{\mathcal{T}_k(X_{c})\}; \theta) - \frac{X_t-X_0}{\sigma_t}\|_2
\end{equation}

\subsubsection{Joint Optimization}
The PTD adopts an end-to-end joint training strategy for both the reconstructive module (PTD$_{\textit{rec}}$) and the conditional diffusion module (PTD$_{\textit{diff}}$), the total loss function is summarized:
\begin{equation}
	\mathcal{L}_{\textit{total}} = \mathcal{L}_{\textit{Content}} + \mathcal{L}_{\textit{Guidance}} + \mathcal{L}_{\textit{Diff}}.
\end{equation}

\section{Experiments and Results}
\subsection{Experiment settings}
\subsubsection{Datasets and implementation details}
We select the ``2016 NIH-AAPM-Mayo Clinic Low-Dose CT Grand Challenge'' dataset \cite{McCollough2016TUFG207A04OO} to evaluate the performance of our method, which contains 5936 slices from 10 patients (8 for training, 1 for validation, and 1 for testing). Following \cite{Xia2024RegFormerAL}, we adopt the distance-driven method with fan-beam geometry to generate sparse-view data, where the distances from the X-ray source to the rotation center and detector are set to 595 mm and 1085.6 mm, respectively. The number of detectors $N_d$ is 368, and the length of each is 2.5716 mm. The height and width of each pixel are both 1.3282 mm. The simulated sparse-view projections are uniformly sampled over the full $360^{\circ}$ range. To simulate the real noise, we also exploit the statistical model of the pre-logarithm projection data:
\begin{equation}
	I = \textit{Poisson}(I_0 \exp(-\hat{y})) + \textit{Normal}(0, \sigma_e^2),
\end{equation}
where $I_0$ is the incident x-ray intensity, $\sigma_e^2$ denotes background electronic noise variance, $\hat{y}$ is noisy-free sinogram data. We set $I_0=1e6, \sigma^2=10$ according to \cite{Xia2024RegFormerAL}. The noisy sinogram data $y$ is calculated by performing the logarithm transformation on the transmission data $I$.

We use a representative algorithm as PTD$_{\textit{rec}}$, i.e., LEARN \cite{chen2018learn}. For PTD$_{\textit{diff}}$, we only adopt the same denoising network as DDPM but using more lightweight configuration, it contains four encoding and four decoding layers, and the feature dimensions are set to $[32, 64, 128, 256]$ in them separately. The whole model is trained on one Nvidia RTX 4090 GPU, and the initial learning rate is set $5e-4$, total iteration is $200k$, and diffusion step is 10 for all the experiments using quadratic discretization. The distortion-aware Peak Signal-to-Noise Ratio (PSNR), Structural Similarity (SSIM), and perception-aware Learned Perceptual Image Patch Similarity (LPIPS) \cite{Zhang2018TheUE} are used as evaluation metrics. The code is released at \href{https://github.com/Wenchao-Du/PTD}{https://github.com/Wenchao-Du/PTD}.

\subsection{Results and Analysis}
\begin{table}[t]
	\centering
	\caption{Comparison of Results on 32 and 64 Projection Views. The best and second best are \textbf{bolded} and \underline{underlined} separately.}
	\setlength{\tabcolsep}{1.5pt}
	\footnotesize
	\begin{tabular}{lcccccc}
		\hline
		\multicolumn{1}{c}{\multirow{2}{*}{\textbf{Method}}} & \multicolumn{3}{c}{\textbf{32-views}} & \multicolumn{3}{c}{\textbf{64-views}} \\
		\cline{2-4} \cline{5-7}
		& PSNR$\uparrow$& SSIM$\uparrow$& LPIPS$\downarrow$(\%) & PSNR$\uparrow$ & SSIM$\uparrow$& LPIPS$\downarrow$(\%) \\
		\hline
		FBP        & 21.08 & 0.307 & 39.97  & 25.22 & 0.489 & 28.56 \\
		FBPConv \cite{jin2017deep} & 33.45 & 0.869  & 6.49  & 38.13 & 0.933 & 3.96 \\
		DuDoTrans \cite{wang2022dudotrans} & 35.26 & 0.895 & 4.79  & 39.52 & 0.950 & 2.55 \\
		LEARN \cite{chen2018learn}     & 39.14 & 0.944 & 2.82  & 42.64 & 0.973 & 1.36 \\
		RegFormer \cite{Xia2024RegFormerAL}  & \textbf{41.17} & \textbf{0.961} & 1.91  & $\underline{44.55}$ & \textbf{0.980} & 0.95 \\
		DDDM \cite{yang2024dual} & 36.53 & 0.907 & 3.42 & -- & -- & -- \\
		\hline
		PTD-1      & $\underline{41.09}$ & $\underline{0.959}$ & 1.84  & \textbf{44.58} & \textbf{0.980} &  0.87 \\
		PTD-5      & 39.97 & 0.950 & 1.29  & 43.69 & $\underline{0.976}$ & $\underline{0.63}$ \\
		PTD-8      & 39.50 & 0.944 & \textbf{1.13} & 43.12 & 0.973 & \textbf{0.51} \\
		\hline
	\end{tabular}
	\label{tab2}
\end{table}
\begin{figure}[t]
	\centering
	\begin{minipage}{1.0\linewidth}
		\includegraphics[width=1.\linewidth]{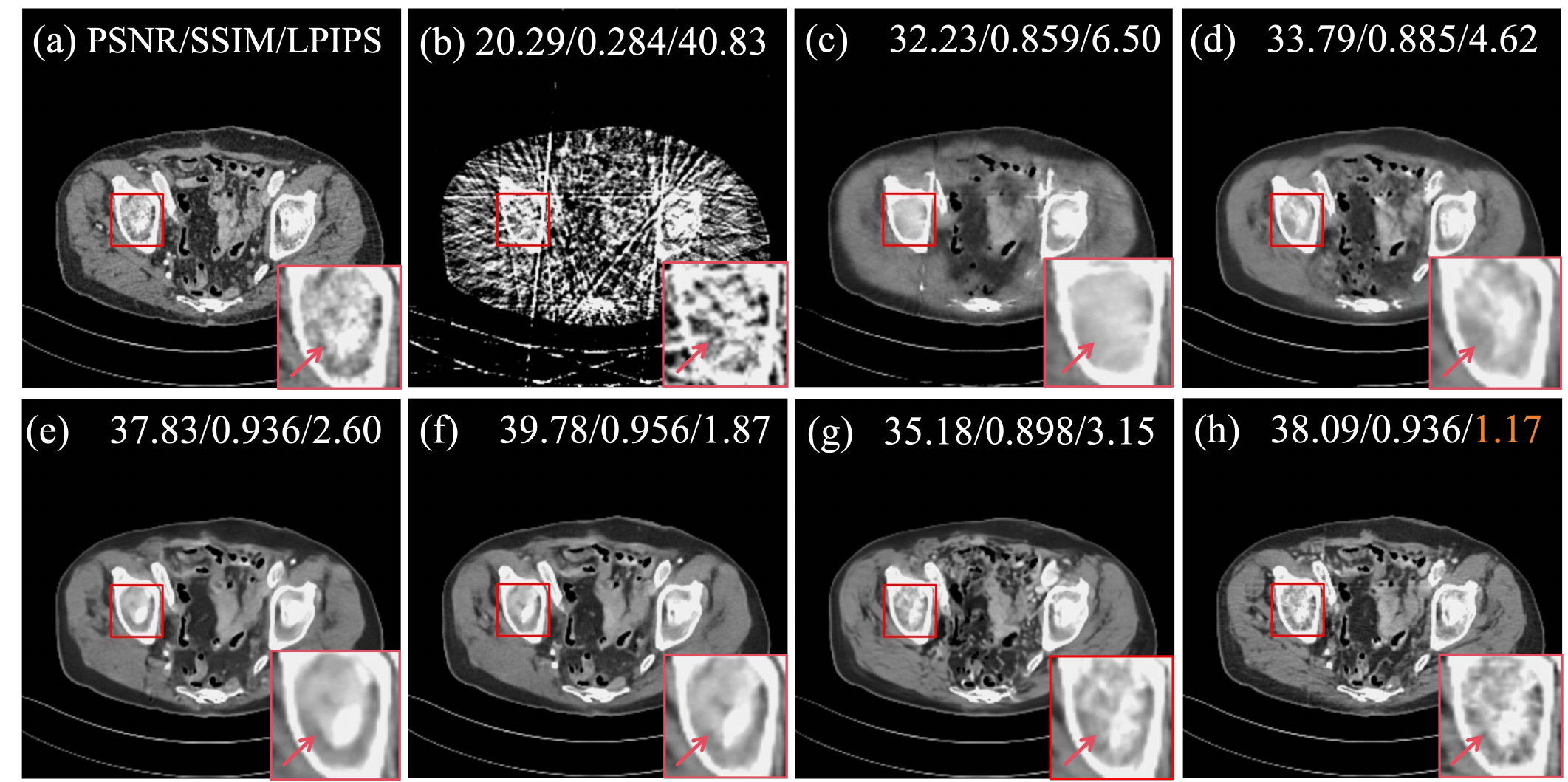}
	\end{minipage}
	\caption{Visualized results on sparse 32-view CT. (a) Ground Truth, (b) FBP, (c) FBPConv, (d) DuDoTrans, (e) LEARN, (f) RegFormer, (g) DDDM and (h) PTD-8. The display window is [-160, 240]HU.}
	\label{fig2}
\end{figure}
\subsubsection{Comparison with SOTA methods}
We select the representative FBPConv \cite{jin2017deep}, DuDoTrans \cite{wang2022dudotrans}, LEARN \cite{chen2018learn}, Regformer \cite{Xia2024RegFormerAL}, and Diffusion-based DDDM \cite{yang2024dual} for comparison. Note that DDDM builds a dual-domain diffusion framework, i.e., projection and image domains diffusion via DDIM model \cite{song2020denoising}. Following official setting, we only evaluate it under 32-view condition. Quantitative results are illuminated in table~\ref{tab2}, all the methods achieve significant performance gains. Specifically, by integrating the non-local image priors into the deep unrolled model, RegFormer has the best PSNR and SSIM values on 32-view CT. Instead, DDDM only achieves slightly better metrics compared to FBPConv and DuDoTrans. Our PTD model presents significant gains against original LEARN, which achieves competitive PSNR and SSIM metrics against RegFormer, and has lower LPIPS value with only one sampling step (denoted by PTD-1). Furthermore, visualized results are shown in Fig~\ref{fig2}. Most methods result in heavy oversmoothness in reconstructed images. DDDM preserves more texture details. However, undesired artifacts and distorted textures also leads to lower PSNR and SSIM values. In contrast, our PTD achieves the sharper details for high-fidelity reconstruction with 8 time steps only (denoted as PTD-8).
\begin{table}[t]
	\centering
	\caption{Ablation studies on each component in our framework.}
	\setlength{\tabcolsep}{2pt}
	\footnotesize
	\begin{tabular}{c|ccc|ccc|c}
		\hline
		\multirow{2}{*}{Task} &
		\multicolumn{3}{c|}{Training Scheme} &
		\multicolumn{3}{c|}{Metric} & \multirow{2}{*}{\#Params (M)}\\
		\cline{2-7}
		& $\mathcal{L}_{\textit{Diff}}$ & $\mathcal{L}_{\textit{Content}}$ & $\mathcal{L}_{\textit{Guidance}}$ & PSNR & SSIM & LPIPS & \\
		\cline{1-8}
		FBP & --& -- & -- & 20.36 & 0.2957  & 39.30 &  -- \\
        \cline{1-8}
		\multirow{3}{*}{SVCT} & $\surd$  &  &  & 35.69 & 0.9031  & 2.192 & 5.19\\
		& $\surd$ & $\surd$ &     & 39.73  & 0.9515 &\textbf{1.040} & 8.58\\
		& $\surd$ & $\surd$ & $\surd$ & \textbf{40.03} & \textbf{0.9541} & 1.066 & 8.64\\
		\hline
	\end{tabular}
	\label{tab3}
\end{table}
\begin{figure}[t]
	\centering
	\begin{minipage}{1.0\linewidth}
		\includegraphics[width=1.\linewidth]{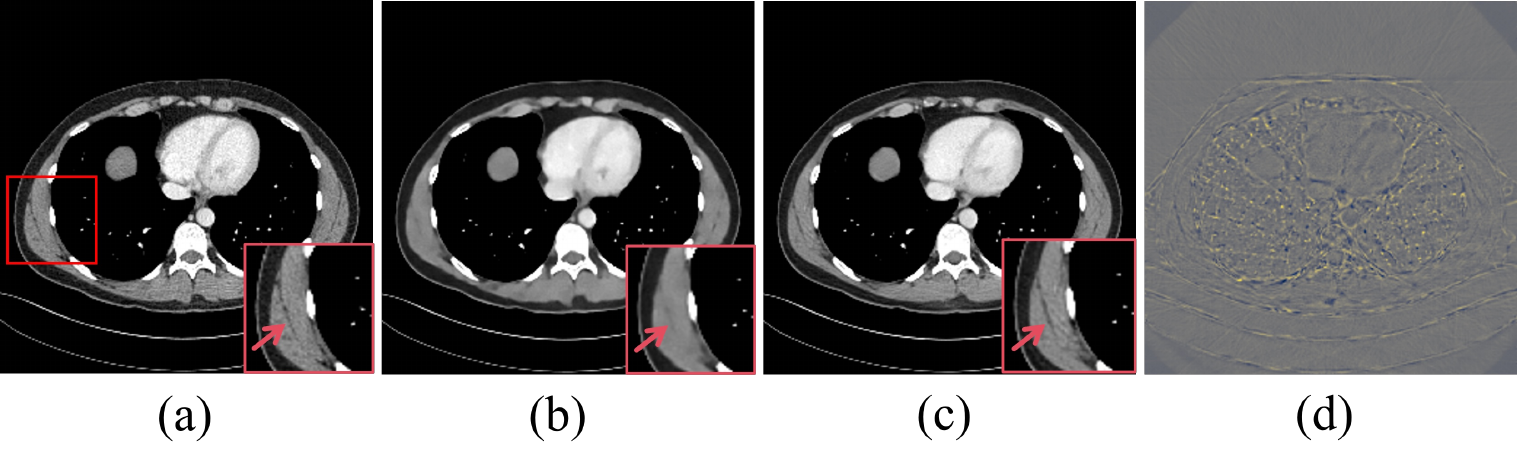}
	\end{minipage}
	\caption{Visualized analysis for key components of our framework on different reconstruction tasks. (a) Ground Truth, (b) Initial reconstruction from basic predictor in Fig~\ref{fig1}, (c) Enhanced Result and (d) Residual map between (b) and (c).}
	\label{fig3}
\end{figure}
\subsubsection{Ablation Studies} We analyze the effects of each component of our PTD, including the reconstruction and conditional guidance modules. Experiments are performed on validation set under sparse 32-view settings. All the variants are trained from scratch with 50k iterations, and sampled with five steps during the inference stage.

Table ~\ref{tab3} gives the detailed results. It is obvious that the DSB model (denoted by $\mathcal{L}_{\textit{Diff}}$) achieves significant performance gains against the initial FBP results, which has been validated in recent advances \cite{gao2023corediff,du2024structure}. We then sequentially insert the reconstructive module and MHFG prompts (denoted by $\mathcal{L}_{\textit{Content}}$ and $\mathcal{L}_{\textit{Guidance}}$) into the baseline, which brings consistent gains in terms of PSNR and SSIM metrics, but the LPIPS values show slight changes. This is because the baseline generates pleasing visual details during the diffusion process, and extra modules lead to the contrast-enhanced structures in the results. Furthermore, inserting the LEARN-based reconstruction module brings more significant gains among metrics. Finally, we insert the MHFG module into the baseline model, which achieves better PSNR and SSIM values.

Fig~\ref{fig3} presents visualized results for key components in our framework. Integrating the initial reconstruction result into the DSB module still generates over-smoothed structures and textures, although it leads to higher PSNR and SSIM metrics. In contrast, our PTD framework can alleviate this problem. As shown in Fig~\ref{fig3}(c), PTD focuses on the contrast-enhanced tissues and reliable texture generation. The residual map (shown in Fig~\ref{fig3}(d)) implies our PTD could bring significantly better structural details and visual appeal.
\section{Conclusion}
This paper proposes a progressively texture-aware diffusion model for sparse-view CT imaging, which integrates a basic reconstructive module and a conditional diffusion model to recover low- and high-frequency image signals in a coarse-to-fine manner. The key to our method is combining effective image content and texture priors to drive deterministic diffusion and produce crucial tissue and structural details, it easily adapts to different low-dose CT tasks with a flexible reconstruction module, and achieves a better trade-off between structural preservation and visual appeal with only a few sampling steps.

\section{Acknowledgment}
This work was supported in part by the National Natural Science Foundation of China under Grants 62301345 and U25A20439, in part by Sichuan Province Postdoctoral Special Funding under Grant TB2025010, in part by the National Basic Scientific Research Project of China under Grant JCKY2024110C080.
\bibliographystyle{IEEEbib}
\bibliography{mybib}

\end{document}